\documentclass[
  journal=medium,
  year=2025,
]{cup-journal}

\usepackage{amsmath}
\usepackage[nopatch]{microtype}
\usepackage{booktabs}
\usepackage{tabularx}
\usepackage{graphicx}
\usepackage{multirow}
\usepackage{comment}
\usepackage[subtle]{savetrees}
\usepackage{enumitem}

\usepackage{arydshln}
\usepackage[table]{xcolor}
\definecolor{verylightgreen}{rgb}{0.93, 1.0, 0.93}
\definecolor{verylightblue}{rgb}{0.90, 0.95, 1.0}
\definecolor{verylightyellow}{rgb}{1.0, 1.0, 0.7}
\definecolor{verylightred}{rgb}{1.0, 0.9, 0.9}
\definecolor{questionbackground}{RGB}{246,255,213}
\definecolor{answerbackground}{RGB}{255,230,213}
\definecolor{cellbackground}{RGB}{213,255,255}
\usepackage{tikz}
\usepackage{array}
\usepackage{booktabs}

\title{Checkmate: interpretable and explainable RSVQA is the endgame}

\author{Lucrezia Tosato\textsuperscript{1,4}}
\author{Christel Tartini Chappuis\textsuperscript{2}}
\author{Syrielle Montariol\textsuperscript{3}}
\author{Flora Weissgerber\textsuperscript{4}}
\author{Sylvain Lobry\textsuperscript{1}}
\author{Devis Tuia\textsuperscript{2}}

\keywords{Remote sensing, visual question answering, interpretability, explainability, deep learning } 
\begin{document}

\begin{flushleft}
\textsuperscript{1}LIPADE, Université Paris Cité, Paris, France \\
\textsuperscript{2}ECEO, Ecole Polytechnique Fédérale de Lausanne, Sion, Switzerland \\
\textsuperscript{3}NLP, Ecole Polytechnique Fédérale de Lausanne, Lausanne, Switzerland \\
\textsuperscript{4}SAPIA, ONERA, Palaiseau, France
\end{flushleft}

\begin{abstract}
Remote Sensing Visual Question Answering (RSVQA) presents unique challenges in ensuring that model decisions are both understandable and grounded in visual content. Current models often suffer from a lack of interpretability and explainability, as well as from biases in dataset distributions that lead to shortcut learning. In this work, we tackle these issues by introducing a novel RSVQA dataset, \textbf{Chessboard}, designed to minimize biases through 3'123'253 questions and a balanced answer distribution. Each answer is linked to one or more cells within the image, enabling fine-grained visual reasoning.

Building on this dataset, we develop an explainable and interpretable model called \textbf{Checkmate} that identifies the image cells most relevant to its decisions. Through extensive experiments across multiple model architectures, we show that our approach improves transparency and supports more trustworthy decision-making in RSVQA systems.
\end{abstract}

\section{Introduction}

Remote Sensing Visual Question Answering (RSVQA)~\cite{lobry2019visual} aims to enable models to answer questions about remote sensing imagery. Two challenges not yet thoroughly explored in this field are the lack of interpretability and explainability in current models, and the presence of biases in existing RSVQA datasets. In the context of RSVQA, biases can lead to models making decisions that are not based on the image content, but rather on answer distributions or the phrasing of questions~\cite{chappuis2023curse}. 
Addressing these issues is important for ensuring that models make accurate, fair, and understandable decisions. 

Interpretability and explainability enable models to make decisions that are not only accurate, but also transparent. Without transparency into how decisions are made, it remains unclear whether models are truly reasoning about visual content or merely exploiting dataset correlations.

Interpretability offers a window into the model’s internal reasoning, while explainability ensures that this reasoning is communicated in a clear and accessible way. Together, they are essential for detecting biases~\cite{gilpin2018explaining} and fostering trust in AI systems~\cite{ferrario2022explainability}. 

The objective of this study is to propose a model that provides interpretable and explainable results, while showcasing the design of a dataset that minimizes biases. To this end, our first contribution is the development of an explainable and interpretable model that is competitive with the state of the art, while being adaptable to other datasets and modalities. Our second contribution is a new RSVQA dataset, named Chessboard, used to support the training and evaluation of our model. Chessboard is designed with biased minimization in mind, 335 possible answers, 519 unique words, and employs a total of 186'731 distinct question templates. The dataset also includes the relative location (within the image) of the elements necessary to answer the question, with up to 16 possible cells per image. Each answer may be grounded in any number of these cells, from none to all of them. Together, these contributions pave the way to trustworthy RSVQA systems. 

\section{Related Work}\label{sec:SoA}
\textbf{RSVQA: }RSVQA extends traditional Visual Question Answering (VQA)~\cite{antol2015vqa}, to the domain of Remote Sensing, providing natural language answers to questions about remote sensing images. The first RSVQA 
pipeline~\cite{lobry2019visual} combines Convolutional Neural Network (CNN)  for images and Recursive Neural Network (RNN) for text features via point-wise multiplication. A significant improvement in visual representation is introduced by~\cite{felix2021cross}, who integrates an object detection module based on Faster R-CNN with BERT for textual input. In contrast,~\cite{zheng2021mutual} focuses on improving the fusion between visual and textual features through a mutual attention mechanism

Recent advancements in the RSVQA architecture include the incorporation of large language models (LLMs), such as in LiT-4-RSVQA, a Transformer-based model proposed by~\cite{hackel2023lit} using Tiny-Bert~\cite{jiao2019tinybert}. Vision-language models (VLMs), such as VisualBERT, have shown promise in joint image-text processing~\cite{siebert2022multi}. Foundation models, as RSChatGPT~\cite{guo2024remote}, further advance the field by integrating task planning and response generation.

To improve data diversity, augmentation techniques are used to enrich the variety of questions. For instance, \cite{yuan2023multilingual} apply multilingual translation, while \cite{boussaid2025llm} use LLMs to rephrase questions.\\
\textbf{Biases in RSVQA: }Despite methodological advancements, RSVQA datasets still suffer from inherent biases. \cite{chappuis2023curse} shows that models often exploit imbalanced answer distributions and question phrasing instead of relying on visual understanding. For example, answer imbalances, such as 52\% "no" in~\cite{lobry2021rsvqa} and 43\% "non flooded" in~\cite{rahnemoonfar2021floodnet}, can lead models to default to dominant responses regardless of the content of the image.\\
\textbf{Interpretability and Explainability in RSVQA: }Interpretability involves designing models that are inherently understandable, while explainability refers to generating post hoc explanations for complex models~\cite{rudin2019stop}. Both are crucial for building trust and for uncovering correlations within the data~\cite{marcinkevivcs2020interpretability}.

A significant portion of existing work in RSVQA does not incorporate interpretability or explainability methods, making it difficult to understand how models reach their predictions or to trust their outputs.
However, more recent efforts have begun to explore ways to improve interpretability. One approach extracts classes from images and feeds them to LLMs for question answering, used in both optical~\cite{chappuis2022prompt} and multi-modal contexts~\cite{tosato2025sar}. Other methods compute attention over segmentation maps~\cite{tosato2024segmentation} or apply graph-based models to capture topological object relations~\cite{zhang2024hierarchical}. Building on these, PAN-RSVQA~\cite{chappuis_pan-rsvqa_2025} uses EO foundation models and multimodal transformers to add interpretable pseudo-annotations.

Explainability remains relatively unexplored in RSVQA, with approaches such as Grad-CAM being used to highlight image regions that influence the model's decision~\cite{zhang2024hierarchical,wang2024rsadapter}. In medical applications, class grounding based on object detection has been investigated~\cite{liu2024gemex}.

To the best of our knowledge, no existing RSVQA model address interpretability and explainability in a unified manner. Moreover, no existing RSVQA model integrates explainability directly into the model in a way that is readily accessible and useful for end users. To fill this gap, we introduce Checkmate, an RSVQA model designed to be both interpretable and explainable. Developed using our newly created Chessboard dataset, which explicitly targets bias mitigation, Checkmate aims to foster trustworthy and transparent decision-making in RSVQA.

\section{Dataset}\label{sec:dataset}

\begin{figure}[t!]
  \centering
  \includegraphics[width=\columnwidth]{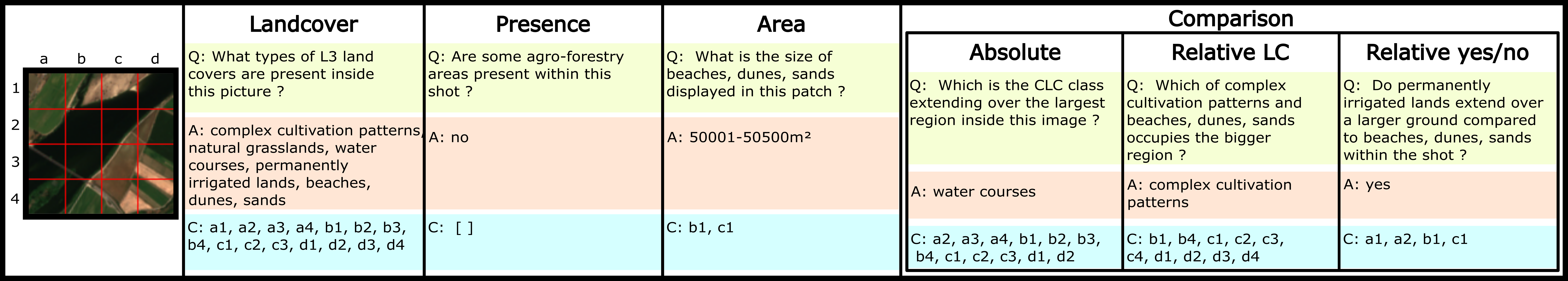}
  \caption{Example of \colorbox{questionbackground}{\textcolor{black}{Question}}, \colorbox{answerbackground}{\textcolor{black}{Answer}}, and \colorbox{cellbackground}{\textcolor{black}{Cells}} in Chessboard for the four question types. The main categories are Landcover, Presence, Area, and Comparison. Within Comparison, there are two subtypes: Absolute and Relative. In Relative Comparison, the answer can be either a land cover class (LC) or a binary yes/no. }
  \label{fig:datasetex}
\end{figure}

\textbf{Images source:} We use images from reBEN~\cite{clasen2024reben}, a large-scale, multi-modal remote sensing dataset. It comprises 549,488 pairs of Sentinel-1 and Sentinel-2 image patches, each of size 120×120 pixels with a spatial resolution of 10 meters, collected from 10 European countries. ReBEN builds on top of BigEarthNet (BEN)~\cite{sumbul2019bigearthnet} by providing pixel-level reference maps for each image patch. These segmentation labels are derived from the most recent CORINE Land Cover (CLC) map of 2018\footnote{https://land.copernicus.eu/content/corine-land-cover-nomenclature-guidelines/html/}. The CLC supports multiple nomenclatures with 3 levels of granularity; for segmentation purposes, reBEN adopts the Level 3 (L3) nomenclature, which includes 44 land cover classes. For the image-level class labels, however, the same 19-class scheme used in BEN is retained. Additionally, reBEN introduces a novel geography-based data split strategy, which substantially reduces spatial correlations across the training, validation, and test sets.\\
\textbf{Questions creation:} We generate four types of questions: \textit{presence}, \textit{land cover}, \textit{area}, and \textit{comparison} (see examples for each type in Figure~\ref{fig:datasetex}). 
Presence questions check if a class is visible; Land cover questions ask which classes are found in the image; Area questions assess the spatial coverage of a class; Comparison questions evaluate differences in size between classes. Comparison questions can be categorized into three subtypes: (i) Relative comparison with a class name answer, where the response identifies a class as being larger or smaller in area than another; (ii) Relative comparison with a yes/no answer, which confirms or denies whether one class covers more or less ground than another; and (iii) Absolute comparison, which assesses the total surface area each land cover class occupies within an image to determine which is the dominant (i.e., most widespread) or minimal (i.e., least prevalent) class overall.
To generate the questions and answers, we use 190'951 templates to ensure high linguistic diversity. A template is a set of pre-defined sentence components. For example, a 4-components template might be: \texttt{{start}}, \texttt{{placeholder}}, 
\texttt{{verb}}, \texttt{{preposition}}, and \texttt{{end}}. A question can have up to 7 components, and each component can be populated by a pre-defined set of words. Words like \textit{Are there} or \textit{Are some} may appear in \texttt{start}, followed by the \texttt{placeholder \#}, while verbs such as \textit{displayed} fill \texttt{verb}, and prepositions such as \textit{within the} or \textit{inside the} complete the structure. All question variations are then automatically generated by combining the words. The \textit{\#} symbol acts as a placeholder for image classes.
For each image, all possible question-answer pairs are generated using randomly selected templates, resulting in 62'353'609 pairs.\\
\textbf{Cells creation: }To incorporate spatial information, the image is divided into 16 cells by creating a 4×4 grid along the x and y axes. We designate these cells using labels for the x-axis (a, b, c, d) and the y-axis (1, 2, 3, 4), resembling the layout of a chessboard (see Figure~\ref{fig:datasetex}). Each cell is 30×30 pixels in size. During the question and answer generation process, the cells in which each class appears are saved. For presence, area and absolute comparison questions, only the class mentioned in the question is considered. For Relative comparison, both classes involved are considered. For land cover questions, all classes present in the image are taken into account. A class is not considered present in a cell if it occupies less than 30 pixels in that cell, and it is not considered present in the image at all if it has fewer than 30 pixels across the entire image.\\
\textbf{Balancing process: }The stochastic process results in an unbalanced dataset; some answers (such as “0m²” and “no”) occur much more frequently than others (“1350001-1355000m²” and “burnt areas”). To solve this problem, a balancing step is carried out differently for each type of question. For presence questions, we ensure an equal number of “yes” and “no” answers per class. The number of examples is defined as the minimum between the occurrences of “yes” and “no” for each class. Land cover questions are all retained, as only one is generated per image. For area questions, the most frequent answers are “0m²”, corresponding to cases where the class is absent, and “1'440'000m²” when the class covers the entire image, which occurs in 93'514 images. Our goal is for “0m²" and “1'440'000m²” to appear as often as the median frequency of the other values. To further improve balance, we group area values into ranges with 5'000m² increments, while still treating “0m²” and “1'440'000m²” as individual categories. For comparison questions, we compute the median number of occurrences across all answers and then cap each answer’s frequency at that median. 
Regarding the cells, no balancing is applied, based on the assumption that land cover classes do not always appear in the same area of an image.

At the end of the process, our dataset comprises 3'123'253 image/question/answer triplets, generated using 186'731 distinct question templates and a vocabulary of 519 unique words, spanning 459'361 images from the original dataset. Our dataset is divided using the same split provided by reBEN, ending up with 49.0\% training set, 27.0\% validation set and 24.0\% test set.
There are 335 possible answers in total, distributed across three categories: 290 correspond to area-related questions, 43 to land cover, and 2 are  “yes”/“no” responses.
 \\
~\vspace{.1em}\\
\textbf{Dataset evaluation: }After the balancing, three metrics from~\cite{chappuis2023curse} are used to evaluate the biases in the dataset: Uniform distribution, Prior distribution, and $L_{B_{\text{score}}}$. The Uniform distribution is defined as $1 / A_{\text{unique}}$, where $A_{\text{unique}}$ is the number of distinct answers. The Prior distribution is calculated as $A_{\text{common}} / N$, with $A_{\text{common}}$ being the count of the most frequent answer and $N$ the total number of samples. The bias score is then given by $L_{B_{\text{score}}} = (\text{Prior} - \text{Uniform}) / (1 - \text{Uniform})$. These scores range from 0 to 1, with lower values indicating less bias. 

In Table~\ref{rsvqabiasscores}, the metrics are applied to the Chessboard, separating the analysis of Answers and Cells, and reporting both average and overall results computed by question type. 
Chessboard has a very low a priori bias, improving the $L_{B_{\text{score}}}$ of All Answer by 83\%, the Average by 87\%, and the Overall by 78\% compared to the datasets analyzed in~\cite{chappuis2023curse}. The bias in the cells is also very low, confirming our assumption that, by default, no cell is significantly more represented than the others.

\begin{table}[b]
\centering
\tiny
\begin{tabular}{|l|c|c|c|c|c|c|c|}
\hline
Question Type & N & $A_{\text{unique}}$ & $A_{\text{common}}$ & Most common & Prior & Uniform & $\mathbf{L_{B_{score}}}$\\
\hline

All Answers & 1'031'798 & 334 & 65'775 & no & 0.06375 & 0.00299 & 0.06094\\
All Cells & 7'108'905 & 16 & 445'357 & b1 & 0.06265 & 0.0625 & 0.00016\\
Average &  &  &  &  & 0.21090 & 0.18134 & 0.03349\\
Overall &  &  &  &  & 0.12968 & 0.07505 & 0.05783 \\
 \hline
\end{tabular}
\caption{Analysis of Chessboard test set using the Prior, Uniform and $L_{B_{score}}$ scores.}
\label{rsvqabiasscores}
\end{table}

\section{Method}\label{sec:method}
\begin{figure}[htbp]
  \centering
  \includegraphics[width=\columnwidth]{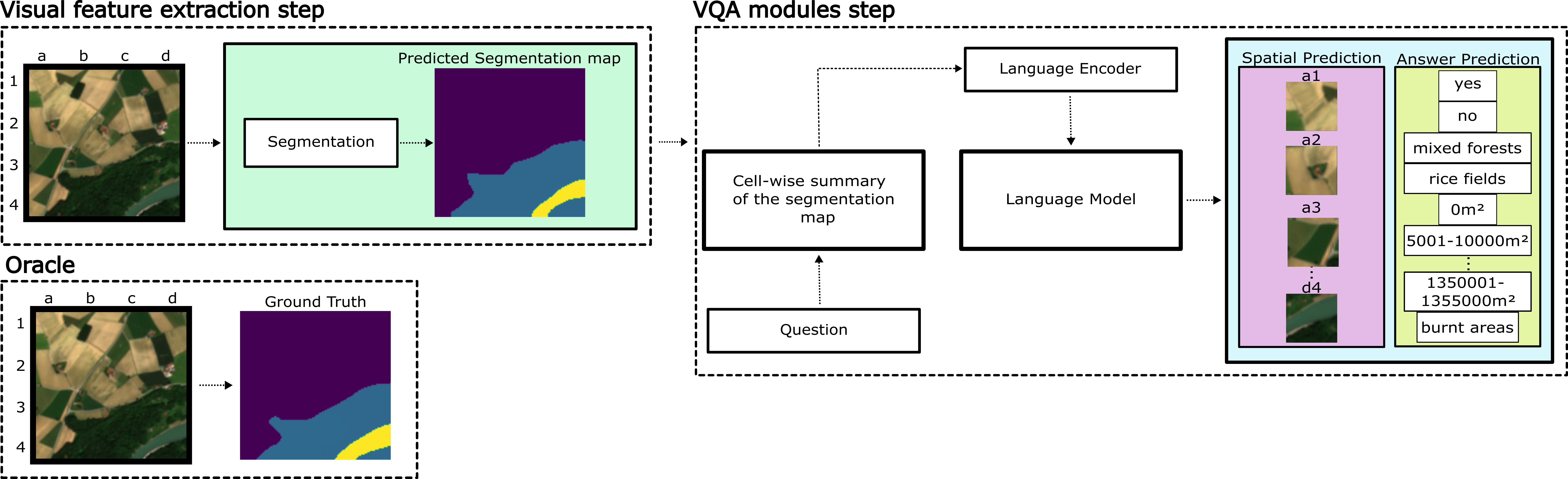}
  \caption{Our proposed method, Checkmate, first transforms the image into a semantic segmentation map, then summarizes it at the cell level, and finally feeds this representation into a language-only model to predict both the answer and the spatial location that influenced the decision.}
  \label{fig:languagemodels}
\end{figure}

In this work we introduce Checkmate, a two-stage model that integrates both interpretability and explainability, ensuring that the decision-making process is transparent and the underlying reasoning is accessible to users. A general overview of our model is shown in Figure~\ref{fig:languagemodels}. The Checkmate model is divided into two main stages: first, visual features are extracted from the image (\textbf{Visual feature extraction step}). These visual features are then translated to text and used to predict the answer and spatial locations that are relevant to the question (\textbf{VQA modules step}). Because of the translation of visual features to text, our model cannot be trained End-to-End. While the Visual feature extraction step can be trained using semantic segmentation ground truth, we use an \textbf{Oracle} model to train the VQA modules step.
\\
\textbf{Visual feature extraction: }
In our approach, the input image is initially processed through a semantic segmentation model. The model predicts the  44 classes present in the images, including an "Unlabeled" category. To achieve the best possible results, we explore several different segmentation models. The first model tested is U-Net, a convolutional neural network specifically designed for semantic segmentation tasks~\cite{ronneberger2015u}. Next, SegFormer is tested~\cite{xie2021segformer}. SegFormer is a transformer-based model that efficiently captures both local and global context. Unlike traditional convolutional approaches, SegFormer combines transformer and convolutional operations in a lightweight, scalable architecture, making it highly effective for segmentation tasks. The last model evaluated is DOFA~\cite{xiong2024neural}, a foundation model inspired by neural plasticity. DOFA is designed to handle various data types and modalities, and has been pre-trained on a large variety of sensors.
\\
\textbf{VQA modules: }The segmentation map is transformed into a textual summary through a structured template. For each cell, the algorithm identifies the present classes. Then, it calculates the number of pixels belonging to each class across the entire image. These values are mapped to one of 290 area ranges as explained in Section~\ref{sec:dataset}. For example, if class \textit{A} appears only in cell a1 with 458 pixels, the generated description regarding this class alone would be: 
 \textit{Table: (a1,  class\_A); Area: class\_A: 45001-50000m²}. 
Since the segmentation maps also include the 'Unlabeled' class, this class is discarded from the summary text.  
This text summary is combined with the question, encoded and fed into a transformer that predicts both the answer and the relevant cells. In this study, we use a DistilBERT~\cite{sanh2019distilbert} model. In detail, we use the last predicted token as an input to two multi-layer perceptrons. The first one has an output layer of 16, corresponding to the different spatial locations in the image. Because several spatial locations can be relevant for a given question, the spatial prediction is treated as a multi-label classification task, and a binary cross-entropy loss is used. For the answer prediction, we use an output layer of dimension 335 and treat the problem as a multi-label classification task, since for land cover questions, multiple classes may be required as the answer. 
Once the answer and the relevant cells have been predicted, a final response is constructed using a predefined template. It begins with "Based on" followed by the predicted cells, then continues with "the answer is" and includes the predicted answer.
If all cells are selected, it will state "all cells", whereas if no cells are selected, it will state "based on the absence of a relevant area."
\\
\textbf{Oracle: } To train the VQA modules and evaluate the model's potential under ideal conditions, we use the ground truth segmentation maps provided in the reBEN dataset. The text summaries generated from these maps are used to train the language model on the answer and cell classification tasks. The model using ground truth segmentation maps is referred to as the Oracle, representing the best possible outcome a semantic segmentation model could achieve. For the models including the Visual feature extraction stage, the weights of the language model obtained from the training with the oracle are reused. This allows for the language model to perform under the assumption of a reliable segmentation and to avoid introducing errors through additional fine-tuning.

\section{Experimental Setup}\label{sec:expe}
\textbf{VQA Modules:} To train the VQA Modules with the Oracle, the summarized text is fed into a DistilBERT~\cite{sanh2019distilbert} model along with the question. Training is performed over 8 epochs with a batch size of 32 and a learning rate of $10^{-5}$. The loss function combines two components: the cell classification loss, weighted by 0.4, and the answer classification loss, weighted by 0.6. This weighting scheme gives slightly more importance to answer classification, which is the primary objective of our VQA pipeline.
\\
\textbf{Semantic Segmentation:} For the segmentation task, various models are evaluated using standard architectures with only the output class count adjusted. All models are trained using cross-entropy loss with a batch size of 32. U-Net and SegFormer are each fully trained for 4 epochs with a learning rate of $10^{-4}$. DOFA is evaluated in two configurations. The first model trains only the decoder with a frozen encoder for 16 epochs and a learning rate of $10^{-4}$. In the second, a new model with the same architecture is initialized, this time with a trainable encoder, and fine-tuned for 3 epochs using a lower learning rate of $10^{-5}$. The decoder in both setups up-samples encoder features using convolutional and bilinear interpolation layers to produce segmentation logits.
\\
\textbf{Baselines:} We compare Checkmate to the following baselines, all treating the VQA task as a classification problem. 

\begin{itemize}
\item The first pipeline, presented by~\cite{lobry2019visual}, uses an \textit{End-to-End} model where image and text features are fused via pointwise multiplication. The visual encoder is a ResNet-50, while the language encoder is DistilBERT. 

\item The second pipeline, named \textit{Image-based} model, is proposed by~\cite{siebert2022multi}. It combines image features extracted by a ResNet-50 with question features from DistilBERT, feeding them into a VisualBERT~\cite{li2019visualbert} architecture.

\item In the \textit{Cell-based} model, we follow a similar logic as in~\cite{su2019vl}, where the image is divided into 16 patches. Each patch is processed through a ResNet-18, and the resulting features are concatenated. These visual features are then combined with question embeddings generated using DistilBERT before being passed through a VisualBERT architecture. 

\item The \textit{PERS} model, introduced by~\cite{he2024pers}, uses separate vision and language encoders with a parameter-efficient vision branch. The encoded features are then fused via a multimodal module leveraging self- and cross-attention to align and integrate visual and textual representations. The visual encoder is a Vision Transformer (ViT)~\cite{dosovitskiy2020image}, and the language encoder is BERT.

\item We conduct an ablation study with a \textit{Blind model}, i.e., a model not using any visual information. The model is trained using only the questions, encoded via BERT~\cite{devlin-etal-2019-bert}
\end{itemize}
All models use a batch size of 32 and the same loss function as the Oracle, ensuring explainability. The \textit{End-to-End} baseline is trained for 5 epochs at a learning rate of $10^{-7}$, while other baselines are trained for 30 epochs at $10^{-6}$.
\\
\textbf{Metrics:} 
To evaluate performance on both semantic segmentation and VQA tasks, we adopt a range of metrics. The semantic segmentation task is evaluated using Pixel Accuracy (PA), Mean Pixel Accuracy (MPA), Mean Intersection over Union (mIoU), Frequency Weighted IoU (FWIoU), as well as Precision (P), Recall (R), and F1 score.. \\
VQA performance is measured using Accuracy (Acc.) per question type, as well as average accuracy (mean across question types) and overall accuracy (total correct predictions over all samples).
Micro F1 is also employed for the land cover (LC) classification task, as it is a multi-class classification problem.  For the evaluation of the cell prediction, the micro F1, Precision and Recall are used. Additionally, self-correlation between predicted cells within an image is used to assess spatial consistency in cell predictions. Some of these metrics are reported using both micro and macro averaging. The micro average computes metrics globally by aggregating the total true positives, false negatives, and false positives across all classes. In contrast, the macro average calculates metrics for each class independently and then computes their unweighted mean, without accounting for class imbalance. 

\section{Results and Discussion}

\textbf{Visual feature extraction (Table~\ref{tab:seg_results}):} We compare four semantic segmentation models: U-Net, SegFormer, DOFA with a frozen encoder, and DOFA fully fine-tuned.
SegFormer consistently outperforms other models, standing out as the most robust model in this comparison. It has the highest micro and macro scores, indicating both overall accuracy and strong handling of rare classes. Its superior mIoU and FWIoU reflect more precise and balanced segmentation across the image. U-Net displays low performance in macro metrics and IoU-based measures, suggesting it may struggle with class imbalance or finer segmentation boundaries. DOFA shows limited performance when not fully trained, particularly in macro recall and F1 score, but exhibits clear improvements once trained on the full dataset, closing the gap with SegFormer. 
\begin{table*}[ht]
\centering
\tiny
\begin{tikzpicture}
\node[draw=verylightgreen, line width=3pt, inner sep=2pt] {
\begin{tabular}{lccccccccccc}
\toprule
\textbf{Model} & \textbf{micro P} & \textbf{micro R} & \textbf{micro F1} & \textbf{macro P} & \textbf{macro R} & \textbf{macro F1} & \textbf{PA} & \textbf{MPA} & \textbf{mIoU} & \textbf{FWIoU} \\
\midrule
U-Net                & 0.601 & 0.601 & 0.601 & 0.309 & 0.277 & 0.279 & 0.601 & 0.284 & 0.198 & 0.437 \\
SegFormer            & \textbf{0.653} & \textbf{0.653} & \textbf{0.653} & \textbf{0.444} & \textbf{0.402} & \textbf{0.399} & \textbf{0.653} & \textbf{0.411} & \textbf{0.284} & \textbf{0.496} \\
DOFA (frozen encoder)  & 0.568 & 0.568 & 0.568 & 0.343 & 0.261 & 0.272 & 0.568 & 0.267 & 0.185 & 0.405 \\
DOFA     & 0.639 & 0.639 & 0.639 & 0.412 & 0.339 & 0.350 & 0.639 & 0.347 & 0.248 & 0.478 \\
\bottomrule
\end{tabular}
};
\end{tikzpicture}
\caption{Performance comparison of semantic segmentation models where P stands for precision, R for recall, PA for pixel accuracy, MPA for mean pixel accuracy, mIoU for mean Intersection over Union, and FWIoU for frequency-weighted IoU. Best results are shown in \textbf{bold}.}
\label{tab:seg_results}
\end{table*}
\\
\textbf{Visual Question Answering (Table~\ref{tab:answer_results}):} 
The Checkmate model using SegFormer as a visual feature extractor achieves the best overall performance, with an average accuracy of 50.9\%, consistently outperforming the others across nearly all metrics. DOFA with a frozen encoder and fully-trained DOFA deliver competitive but lower results, while U-Net falls behind in all metrics. This ranking mirrors the segmentation task results, highlighting that better segmentation quality directly contributes to improved VQA performance.

The second part of the table, which compares Checkmate model to alternative architectures, confirms the best performance for our approach. Among the competing methods, the Cell-based model also performs well, coming close to SegFormer in several metrics; making it a good, although non-interpretable, alternative for the answer task. 

\begin{table*}[ht]
\centering
\tiny
\begin{tikzpicture}
\node[draw=verylightyellow, line width=3pt, inner sep=2pt] {
\begin{tabular}{lccccccc}
\toprule
\textbf{Model} & \textbf{Acc. presence} & \textbf{Acc. comparison} & \textbf{Acc. LC} & 
\textbf{micro F1 LC}& 
\textbf{Acc. area} & \textbf{Overall Acc.} & \textbf{Average. Acc.}  \\
\midrule
\multicolumn{8}{c}{Checkmate} \\
\midrule
U-Net                &  0.748 & 0.657 & 0.664 & 0.712 & 0.016 & 0.341 & 0.477 \\
SegFormer           & 0.800 & \textbf{0.696} & \textbf{0.705} &  \textbf{0.747} &0.022 & \textbf{0.364} & \textbf{0.509}  \\
DOFA (frozen encoder)         & 0.742 &  0.643 & 0.651 & 0.691 & 0.014 & 0.334 & 0.469 \\
DOFA   & 0.766 & 0.678 & 0.692 & 0.733 & 0.017 & 0.353 & 0.491 \\
\hdashline
Oracle               & 0.999 & 0.944 & 0.999 & 0.999 & 0.831 & 0.901 & 0.943 \\
\midrule
\multicolumn{8}{c}{Baselines} \\
\midrule
End to end           & 0.498 & 0.159 & 0.455 & 0.480 & 0.011 & 0.164 & 0.321 \\
Image-based          & 0.780 & 0.630 & 0.618 & 0.651 & 0.022 & 0.334 & 0.474 \\
Cell-based           & \textbf{0.821} & 0.671 & 0.663 & 0.697 & \textbf{0.024} & 0.355 & 0.502 \\
PERS                 & 0.786 & 0.544 & 0.619 & 0.655 &  0.021 & 0.314 & 0.475 \\
Blind model          & 0.459 & 0.601 & 0.455 & 0.481 & 0.021 & 0.268 & 0.312 \\
\bottomrule
\end{tabular}
};
\end{tikzpicture}
\caption{Performance comparison per question type between Checkmate and the baselines for the answer prediction task. Acc stands for accuracy, LC for land cover class. The best result (excluding Oracle) is shown in \textbf{bold}.}
\label{tab:answer_results}
\end{table*}
\noindent\textbf{Cell Prediction (Table~\ref{tab:cells_results}):} The Cell-based model achieves the highest F1 score and Recall, demonstrating strong true positive detection. However, its lower Precision and high correlation suggest overprediction, leading to more false positives and reduced localization sensitivity. Our proposed model using SegFormer, by contrast, achieves the highest Precision and lowest correlation, producing more diverse and focused predictions. This indicates better discrimination across cells and less overfitting, though it slightly underperforms in Recall, possibly missing some true positives.

\begin{figure}[htbp]
  \centering
  \includegraphics[width=\columnwidth]{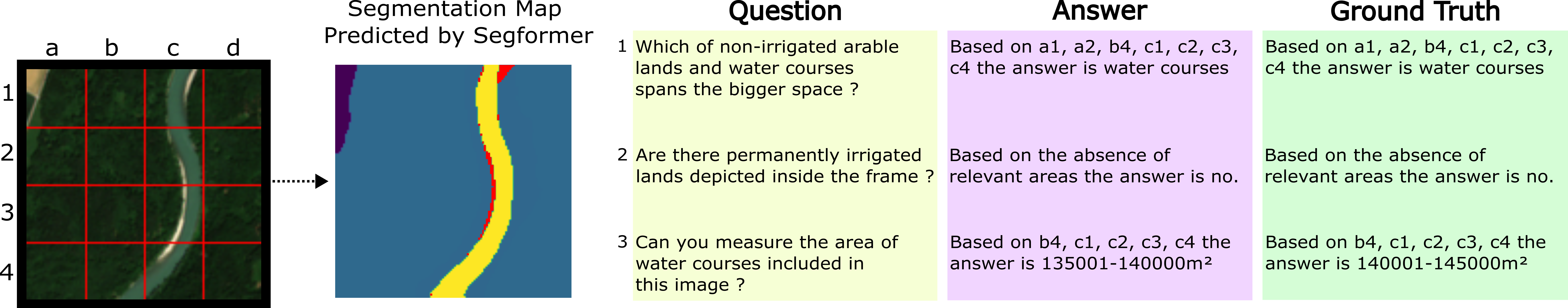}
  \caption{Checkmate outputs using SegFormer with input image, predicted segmentation, sample QA pairs, and ground truth. The red in the map highlights missing pixels of the yellow water course class (in yellow).}
  \label{fig:visualmodels}
\end{figure}

\begin{table*}[ht]
\centering
\tiny
\begin{tikzpicture}
\node[draw=verylightred, line width=3pt, inner sep=2pt] {
\begin{tabular}{lcccc}
\toprule
\textbf{Model} &  \textbf{micro F1} & \textbf{Precision} & \textbf{Recall} & \textbf{Correlation}  \\
\midrule
\multicolumn{5}{c}{Checkmate} \\
\midrule
U-Net                 & 0.796 &  0.865 & 0.738 &  0.670 \\
SegFormer          & 0.825 &  \textbf{0.877} & 0.780 & \textbf{0.639} \\
DOFA (frozen encoder)             & 0.788 &  0.846 & 0.736 & 0.666\\
DOFA     & 0.813 &  0.871 & 0.761 & 0.667 \\
\hdashline 
Oracle               & 0.999 & 0.999 & 0.999 & 0.475 \\
\midrule
\multicolumn{5}{c}{Comparison with other models} \\
\midrule
End to end          & 0.764  & 0.646 & 0.933 & 0.986 \\
Image-based         & 0.806 &  0.736 & 0.891 & 0.847 \\
Cell-based          & \textbf{0.829} & 0.770 &  \textbf{0.897} & 0.734 \\
PERS                & 0.800 & 0.728 &  0.886 & 0.908 \\
Blind model         & 0.764 & 0.646 & 0.933 & 0.993  \\
\bottomrule
\end{tabular}
};
\end{tikzpicture}
\caption{Performance comparison across different models for the cell prediction task. The best result (excluding the Oracle) is indicated in \textbf{bold}.}
\label{tab:cells_results}
\end{table*}
Figure~\ref{fig:visualmodels} shows a sample image, its SegFormer generated segmentation map, and example questions with predicted and ground truth answers. Thanks to the presence of the cells in the answers, it becomes easier to understand whether the model is actually understanding the content of the image.
In the first question, which involves a comparison between two classes, the answer includes cells corresponding to both classes, indicating that the model is able to identify where these classes are located in the image.
This is even more evident in Question 3, which is about the area of water courses. In this case, the model correctly identifies only the cells related to the water course. Although the final answer is incorrect, the segmentation map reveals that this is due to 53 pixels (shown in red) not being predicted. As a result, the model selects a slightly smaller, adjacent range rather than the correct one.
Notably, the mode of the distribution is one, meaning that the most missed predictions are just one range away from the correct one, and close to 20\% are within five ranges of it. 
Similarly, in the second presence-type question, the model answers 'no', correctly identifying the absence of relevant cells and showing it understands the class is not present.

\section{Conclusion and Limitations}
\label{sec:conclusion}
In this work, we tackle key challenges in Remote Sensing Visual Question Answering (RSVQA), particularly the limited interpretability of existing models and the presence of dataset biases that often lead to shortcut learning. To address these issues, we introduce Chessboard, a large-scale and the most balanced RSVQA dataset to date designed to support fine-grained visual reasoning. By linking each answer to specific image cells, Chessboard encourages models to ground their predictions in the visual content.

Building on the structure of the dataset, we introduce Checkmate, a Segmentation-Based RSVQA model which enhances both explainability and interpretability by explicitly identifying the image regions most relevant to each answer using the predicted segmentation maps. We evaluate three segmentation strategies: a CNN-based model (U-Net), a transformer-based model (SegFormer), and a foundation model (DOFA). Among these, SegFormer achieves the best performances, with an F1-score of 65.3\%. Overall, our model reaches an average answer accuracy of 50.9\% and demonstrates superior transparency compared to traditional methods, as reflected by a high F1-micro score of 82.5\% in cell prediction and a lower inter-cell correlation, indicating reduced redundancy and more focused visual reasoning. Despite promising results, our approach exhibits limitations in predicting area-related answers, likely due to the high number of area classes (290).

Overall, our contributions represent a significant step toward more interpretable and trustworthy RSVQA systems, enabling more reliable human-AI collaboration in remote sensing. Future work may involve applying Checkmate to different datasets that provide segmentation maps, such as~\cite{boussaidtosato_tammy_2025,xia2025openearthmap} or modalities and exploring ways to improve performance on area-related tasks.
\newpage



\printbibliography[]

@String(ICCV= {Int. Conf. Comput. Vis.})

@String(CVPRW= {IEEE Conf. Comput. Vis. Pattern Recog. Worksh.})

@String(ICCV  = {ICCV})

@String(CVPRW= {CVPRW})

@inproceedings{ferrario2022explainability,
  title={How explainability contributes to trust in AI},
  author={Ferrario, Andrea and Loi, Michele},
  booktitle={Proceedings of the 2022 ACM conference on fairness, accountability, and transparency},
  pages={1457--1466},
  year={2022}
}

@inproceedings{gilpin2018explaining,
  title={Explaining explanations: An overview of interpretability of machine learning},
  author={Gilpin, Leilani H and Bau, David and Yuan, Ben Z and Bajwa, Ayesha and Specter, Michael and Kagal, Lalana},
  booktitle={2018 IEEE 5th International Conference on data science and advanced analytics (DSAA)},
  pages={80--89},
  year={2018},
  organization={IEEE}
}

@article{chappuis2023curse,
	title = {Evaluating {Language} {Biases} in {Remote} {Sensing} {VQA}},
	journal = {accepted for publication in the IEEE Geoscience and Remote Sensing Magazine (GRSM)},
	author = {Chappuis, Christel and Walt, Eliot and Mendez, Vincent and Lobry, Sylvain and Le Saux, Bertrand and Tuia, Devis},
	year = {2025},
}

@inproceedings{antol2015vqa,
  title={{VQA: Visual Question Answering}},
  author={Antol, Stanislaw and Agrawal, Aishwarya and Lu, Jiasen and Mitchell, Margaret and Batra, Dhruv and Zitnick, C Lawrence and Parikh, Devi},
  booktitle={IEEE/CVF ICCV},
  pages={2425--2433},
  year={2015}
}

@inproceedings{lobry2019visual,
  title={Visual question answering from remote sensing images},
  author={Lobry, Sylvain and Murray, Jesse and Marcos, Diego and Tuia, Devis},
  booktitle={IGARSS 2019-2019 IEEE International Geoscience and Remote Sensing Symposium},
  pages={4951--4954},
  year={2019},
  organization={IEEE}
}

@inproceedings{hackel2023lit,
  title={Lit-4-rsvqa: Lightweight transformer-based visual question answering in remote sensing},
  author={Hackel, Leonard and Clasen, Kai Norman and Ravanbakhsh, Mahdyar and Demir, Beg{\"u}m},
  booktitle={IGARSS 2023-2023 IEEE International Geoscience and Remote Sensing Symposium},
  pages={2231--2234},
  year={2023},
  organization={IEEE}
}

@inproceedings{felix2021cross,
  title={Cross-modal visual question answering for remote sensing data: The international conference on digital image computing: Techniques and applications (DICTA 2021)},
  author={Felix, Rafael and Repasky, Boris and Hodge, Samuel and Zolfaghari, Reza and Abbasnejad, Ehsan and Sherrah, Jamie},
  booktitle={2021 Digital Image Computing: Techniques and Applications (DICTA)},
  pages={1--9},
  year={2021},
  organization={IEEE}
}

@inproceedings{devlin-etal-2019-bert,
    title = "{BERT}: Pre-training of Deep Bidirectional Transformers for Language Understanding",
    author = "Devlin, Jacob  and
      Chang, Ming-Wei  and
      Lee, Kenton  and
      Toutanova, Kristina",
    editor = "Burstein, Jill  and
      Doran, Christy  and
      Solorio, Thamar",
    booktitle = "Proceedings of the 2019 Conference of the North {A}merican Chapter of the Association for Computational Linguistics: Human Language Technologies, Volume 1 (Long and Short Papers)",
    month = jun,
    year = "2019",
    address = "Minneapolis, Minnesota",
    publisher = "Association for Computational Linguistics",
    url = "https://aclanthology.org/N19-1423/",
    doi = "10.18653/v1/N19-1423",
    pages = "4171--4186"
}

@inproceedings{yuan2023multilingual,
  title={Multilingual augmentation for robust visual question answering in remote sensing images},
  author={Yuan, Zhenghang and Mou, Lichao and Zhu, Xiao Xiang},
  booktitle={2023 Joint Urban Remote Sensing Event (JURSE)},
  pages={1--4},
  year={2023},
  organization={IEEE}
}

@inproceedings{siebert2022multi,
  title={Multi-modal fusion transformer for visual question answering in remote sensing},
  author={Siebert, Tim and Clasen, Kai Norman and Ravanbakhsh, Mahdyar and Demir, Beg{\"u}m},
  booktitle={Image and Signal Processing for Remote Sensing XXVIII},
  volume={12267},
  pages={162--170},
  year={2022},
  organization={SPIE}
}

@article{zheng2021mutual,
  title={Mutual attention inception network for remote sensing visual question answering},
  author={Zheng, Xiangtao and Wang, Binqiang and Du, Xingqian and Lu, Xiaoqiang},
  journal={IEEE Transactions on Geoscience and Remote Sensing},
  volume={60},
  pages={1--14},
  year={2021},
  publisher={IEEE}
}

@inproceedings{guo2024remote,
  title={Remote sensing chatgpt: Solving remote sensing tasks with chatgpt and visual models},
  author={Guo, Haonan and Su, Xin and Wu, Chen and Du, Bo and Zhang, Liangpei and Li, Deren},
  booktitle={IGARSS 2024-2024 IEEE International Geoscience and Remote Sensing Symposium},
  pages={11474--11478},
  year={2024},
  organization={IEEE}
}

@inproceedings{tosato2024segmentation,
  title={Segmentation-guided attention for visual question answering from remote sensing images},
  author={Tosato, Lucrezia and Boussaid, Hichem and Weissgerber, Flora and Kurtz, Camille and Wendling, Laurent and Lobry, Sylvain},
  booktitle={IGARSS 2024-2024 IEEE International Geoscience and Remote Sensing Symposium},
  pages={2750--2754},
  year={2024},
  organization={IEEE}
}

@inproceedings{chappuis2022prompt,
  title={Prompt-RSVQA: Prompting visual context to a language model for remote sensing visual question answering},
  author={Chappuis, Christel and Zermatten, Val{\'e}rie and Lobry, Sylvain and Le Saux, Bertrand and Tuia, Devis},
  booktitle={Proceedings of the IEEE/CVF conference on computer vision and pattern recognition},
  pages={1372--1381},
  year={2022}
}

@article{marcinkevivcs2020interpretability,
  title={Interpretability and explainability: A machine learning zoo mini-tour},
  author={Marcinkevi{\v{c}}s, Ri{\v{c}}ards and Vogt, Julia E},
  journal={arXiv preprint arXiv:2012.01805},
  year={2020}
}

@article{rudin2019stop,
  title={Stop explaining black box machine learning models for high stakes decisions and use interpretable models instead},
  author={Rudin, Cynthia},
  journal={Nature machine intelligence},
  volume={1},
  number={5},
  pages={206--215},
  year={2019},
  publisher={Nature Publishing Group UK London}
}

@article{zhang2024hierarchical,
  title={Hierarchical Multi-Modality Graph Reasoning for Remote Sensing Visual Question Answering},
  author={Zhang, Han and Wang, Keming and Zhang, Laixian and Wang, Bingshu and Li, Xuelong},
  journal={IEEE Transactions on Geoscience and Remote Sensing},
  year={2024},
  publisher={IEEE}
}

@article{wang2024rsadapter,
  title={Rsadapter: Adapting multimodal models for remote sensing visual question answering},
  author={Wang, Yuduo and Ghamisi, Pedram},
  journal={IEEE Transactions on Geoscience and Remote Sensing},
  year={2024},
  publisher={IEEE}
}

@article{liu2024gemex,
  title={GEMeX: A Large-Scale, Groundable, and Explainable Medical VQA Benchmark for Chest X-ray Diagnosis},
  author={Liu, Bo and Zou, Ke and Zhan, Liming and Lu, Zexin and Dong, Xiaoyu and Chen, Yidi and Xie, Chengqiang and Cao, Jiannong and Wu, Xiao-Ming and Fu, Huazhu},
  journal={arXiv preprint arXiv:2411.16778},
  year={2024}
}

@article{clasen2024reben,
  title={reben: Refined bigearthnet dataset for remote sensing image analysis},
  author={Clasen, Kai Norman and Hackel, Leonard and Burgert, Tom and Sumbul, Gencer and Demir, Beg{\"u}m and Markl, Volker},
  journal={arXiv preprint arXiv:2407.03653},
  year={2024}
}

@inproceedings{sumbul2019bigearthnet,
  title={Bigearthnet: A large-scale benchmark archive for remote sensing image understanding},
  author={Sumbul, Gencer and Charfuelan, Marcela and Demir, Beg{\"u}m and Markl, Volker},
  booktitle={IGARSS 2019-2019 IEEE international geoscience and remote sensing symposium},
  pages={5901--5904},
  year={2019},
  organization={IEEE}
}

@inproceedings{ronneberger2015u,
  title={{U-NET}: Convolutional networks for biomedical image segmentation},
  author={Ronneberger, Olaf and Fischer, Philipp and Brox, Thomas},
  booktitle={Medical image computing and computer-assisted intervention--MICCAI 2015: 18th international conference, Munich, Germany, October 5-9, 2015, proceedings, part III 18},
  pages={234--241},
  year={2015},
  organization={Springer}
}

@article{xie2021segformer,
  title={SegFormer: Simple and efficient design for semantic segmentation with transformers},
  author={Xie, Enze and Wang, Wenhai and Yu, Zhiding and Anandkumar, Anima and Alvarez, Jose M and Luo, Ping},
  journal={Advances in neural information processing systems},
  volume={34},
  pages={12077--12090},
  year={2021}
}

@article{xiong2024neural,
  title={Neural plasticity-inspired foundation model for observing the Earth crossing modalities},
  author={Xiong, Zhitong and Wang, Yi and Zhang, Fahong and Stewart, Adam J and Hanna, Jo{\"e}lle and Borth, Damian and Papoutsis, Ioannis and Le Saux, Bertrand and Camps-Valls, Gustau and Zhu, Xiao Xiang},
  journal={arXiv e-prints},
  pages={arXiv--2403},
  year={2024}
}

@article{su2019vl,
  title={Vl-bert: Pre-training of generic visual-linguistic representations},
  author={Su, Weijie and Zhu, Xizhou and Cao, Yue and Li, Bin and Lu, Lewei and Wei, Furu and Dai, Jifeng},
  journal={arXiv preprint arXiv:1908.08530},
  year={2019}
}

@article{he2024pers,
  title={PERS: Parameter-Efficient Multi-modal Transfer Learning for Remote Sensing Visual Question Answering},
  author={He, Jinlong and Liu, Gang and Li, Pengfei and Su, Xiaonan and Jiang, Wenhua and Zhang, Dongze and Zhong, Shenjun},
  journal={IEEE Journal of Selected Topics in Applied Earth Observations and Remote Sensing},
  year={2024},
  publisher={IEEE}
}

@inproceedings{chappuis_pan-rsvqa_2025,
	address = {Nashville, TN, USA},
	title = {{PAN}-{RSVQA}: {Vision} {Foundation} {Models} as {Pseudo}-{ANnotators} for {Remote} {Sensing} {Visual} {Question} {Answering}},
	shorttitle = {{PAN}-{RSVQA}},
	language = {en},
	booktitle = {Proceedings of the 2025 {IEEE}/{CVF} {Conference} on {Computer} {Vision} and {Pattern} {Recognition} {Workshops} ({CVPRW})},
	publisher = {IEEE},
	author = {Chappuis, Christel and Sümbül, Gencer and Montariol, Syrielle and Lobry, Sylvain and Tuia, Devis},
	year = {2025},
}

@article{sanh2019distilbert,
  title={DistilBERT, a distilled version of BERT: smaller, faster, cheaper and lighter},
  author={Sanh, Victor and Debut, Lysandre and Chaumond, Julien and Wolf, Thomas},
  journal={arXiv preprint arXiv:1910.01108},
  year={2019}
}

@article{li2019visualbert,
  title={Visualbert: A simple and performant baseline for vision and language},
  author={Li, Liunian Harold and Yatskar, Mark and Yin, Da and Hsieh, Cho-Jui and Chang, Kai-Wei},
  journal={arXiv preprint arXiv:1908.03557},
  year={2019}
}

@article{dosovitskiy2020image,
  title={An image is worth 16x16 words: Transformers for image recognition at scale},
  author={Dosovitskiy, Alexey and Beyer, Lucas and Kolesnikov, Alexander and Weissenborn, Dirk and Zhai, Xiaohua and Unterthiner, Thomas and Dehghani, Mostafa and Minderer, Matthias and Heigold, Georg and Gelly, Sylvain and others},
  journal={arXiv preprint arXiv:2010.11929},
  year={2020}
}

@inproceedings{boussaidtosato_tammy_2025,
	address = {Nashville, TN, USA},
	title = {Visual Question Answering on Multiple Remote Sensing Image Modalities},
	shorttitle = {},
	language = {en},
	booktitle = {Proceedings of the 2025 {IEEE}/{CVF} {Conference} on {Computer} {Vision} and {Pattern} {Recognition} {Workshops} ({CVPRW})},
	publisher = {IEEE},
	author = { Boussaid, Hichem and Tosato, Lucrezia and Weissgerber, Flora and Kurtz, Camille and Wendling, Laurent and Lobry, Sylvain},
	year = {2025},
}

@article{xia2025openearthmap,
  title={OpenEarthMap-SAR: A Benchmark Synthetic Aperture Radar Dataset for Global High-Resolution Land Cover Mapping},
  author={Xia, Junshi and Chen, Hongruixuan and Broni-Bediako, Clifford and Wei, Yimin and Song, Jian and Yokoya, Naoto},
  journal={arXiv preprint arXiv:2501.10891},
  year={2025}
}

@inproceedings{boussaid2025llm,
  title={{LLM}-Driven Data Augmentation for Visual Question Answering},
  author={Hichem, Boussaid and Kwon, Nayoung and Kurtz, Camille and Wendling, Laurent and Lobry, Sylvain},
  booktitle={2025 Joint Urban Remote Sensing Event (JURSE)},
  pages={1--4},
  year={2025},
  organization={IEEE}
}

@article{jiao2019tinybert,
  title={Tinybert: Distilling bert for natural language understanding},
  author={Jiao, Xiaoqi and Yin, Yichun and Shang, Lifeng and Jiang, Xin and Chen, Xiao and Li, Linlin and Wang, Fang and Liu, Qun},
  journal={arXiv preprint arXiv:1909.10351},
  year={2019}
}

@inproceedings{lobry2021rsvqa,
  title={RSVQA meets BigEarthNet: a new, large-scale, visual question answering dataset for remote sensing},
  author={Lobry, Sylvain and Demir, Beg{\"u}m and Tuia, Devis},
  booktitle={2021 IEEE International Geoscience and Remote Sensing Symposium IGARSS},
  pages={1218--1221},
  year={2021},
  organization={IEEE}
}

@article{rahnemoonfar2021floodnet,
  title={Floodnet: A high resolution aerial imagery dataset for post flood scene understanding},
  author={Rahnemoonfar, Maryam and Chowdhury, Tashnim and Sarkar, Argho and Varshney, Debvrat and Yari, Masoud and Murphy, Robin Roberson},
  journal={IEEE Access},
  volume={9},
  pages={89644--89654},
  year={2021},
  publisher={IEEE}
}

@article{tosato2025sar,
  title={SAR Strikes Back: A New Hope for RSVQA},
  author={Tosato, Lucrezia and Lobry, Sylvain and Weissgerber, Flora and Wendling, Laurent},
  journal={IEEE Journal of Selected Topics in Applied Earth Observations and Remote Sensing},
  year={2025},
  publisher={IEEE}
}
\appendix

\end{document}